\documentclass[11pt,a4paper]{article}

\usepackage[T1]{fontenc}
\usepackage[utf8]{inputenc}
\usepackage{tipa}
\usepackage{amsmath,amssymb,amsfonts}
\usepackage{graphicx}
\usepackage{booktabs}
\usepackage{hyperref}
\usepackage{multirow}
\usepackage{cite}
\usepackage{algorithm}
\usepackage{algpseudocode}
\usepackage[margin=1in]{geometry}
\usepackage{microtype}
\usepackage{times}

\title{Arithmetic Pedagogy for Language Models}

\author{
  Andhika Bernard Lumbantobing\footnotemark[1]
  \and
  Hokky Situngkir\footnotemark[2]}
  
\date{}

\begin{document}

\maketitle

\begin{abstract}
We investigate whether methods of human mathematics pedagogy can guide the
training of language models toward arithmetic reasoning. Building on the
GASING method---an Indonesian pedagogy that solves basic arithmetic through a
left-to-right procedure aligned with the causal order of token generation---we
operationalize each operation as a computational procedure whose execution
trace is serialized into natural-language Chain-of-Thought (CoT) supervision.
A small GPT-2 decoder (86M parameters) with a syllabic-agglutinative TOBA
tokenizer for Indonesian is trained from scratch on this data using only a
next-token prediction objective, without reinforcement learning or
reward-based optimization. Monitoring training reveals three distinct learning
phases, and mechanistic analyses---attention-masking interventions on the CoT
information graph, residual-stream probing, and logit-lens inspection---show
that the model first internalizes a procedural pathway 
and subsequently develops an associative, ``mental-arithmetic'' capacity that
retrieves intermediate results without explicit step-by-step computation. The
trained model reaches over 80\% accuracy on held-out problems and attains
competitive performance against substantially larger language models,
indicating that targeted, pedagogically grounded training can yield strong and
economical arithmetic capability at small scale.
\end{abstract}


{
  \renewcommand{\thefootnote}{\fnsymbol{footnote}}
  \footnotetext[1]{Bandung Fe Institute \& Adjunct Science Fellow in InaAI, \texttt{nad@compsoc.bandungfe.net}}
  \footnotetext[2]{AI Research Center IT Del \& Bandung Fe Institute, \texttt{hokky.situngkir@del.ac.id}}
}

\section{Introduction}
	
In the early stages of their development, Transformer-based language
models exhibited rather limited capabilities in mathematical
computation, which was initially often interpreted as little more than a
form of imitation and statistical pattern matching over the massive data
used in their training. Alongside this ran a line of criticism that
positioned large language models (LLMs) as ``\emph{stochastic parrots}'':
capable of producing seemingly meaningful sequences of language through
probabilistic modeling of text distributions, yet not necessarily
possessing the semantic grounding or stable conceptual understanding
characteristic of humans \cite{bender2021}. Subsequent studies then
showed that a number of capabilities, including arithmetic reasoning,
improved spontaneously as model scale increased, whether in terms of the
number of parameters, the volume of data, or the training compute, a
phenomenon discussed as ``\emph{emergent abilities}''
\cite{wei2022b}, although the status of this ``\emph{emergence}'' remains
a matter of debate \cite{schaeffer2023,krakauer2026}.
Through large-scale training, Transformer models appear to form internal
representations that approximate part of the structural regularity
in mathematical relationships by capturing statistical regularities
associated with operations, numerical relations, and problem-solving
patterns \cite{li2022,charton2021}.

To evaluate these capabilities, a number of benchmarks subsequently
emerged \cite{cobbe2021}, containing large volumes of arithmetic
problems specifically designed to test the step-by-step
problem-solving ability of language models. The results of these
evaluations then revealed that model performance can improve
substantially when the model is guided to perform ``reasoning''
by generating intermediate steps explicitly before producing the final
answer \cite{wei2022b,kojima2022}.
This finding has inevitably spurred the development of various
language-model engineering techniques oriented toward enhancing
reasoning capabilities
\cite{yao2023,lightman2023,shao2024}.
Even so, relatively few studies have explored the pedagogical
dimension, namely the extent to which learning methods designed to
build mathematical understanding in humans can be adapted to guide
the training process of language models \cite{gunasekar2023}.
This question is important because mathematics pedagogy not only
arranges the sequence of material but also shapes a conceptual
trajectory: how a system learns to recognize patterns of
numeracy, to link concrete representations with formal symbols, and to
develop problem-solving procedures. If there exist pedagogical
approaches proven effective in improving computational skill and
mathematical understanding in humans, then the question is whether
those pedagogical principles can likewise enhance the performance of
language models when implemented as schemes of data, supervision, or
inference procedures.

In this study, we explore how methods of mathematics pedagogy
can be applied to the training of small-scale Transformer models,
focusing on the learning of basic arithmetic as the primary domain of
observation. Small-scale experiments in a controlled environment,
as conducted in prior work \cite{lee2023},
offer a more transparent path of analysis than observing
frontier models that are more complex or closed. With this
working setting, we are able to observe the emergence trajectory of
reasoning capability and the changes in internal representations during
the training process \cite{power2022,nanda2023}, as well as to
apply behavioral engineering through computational interventions
on the model \cite{saha2025}. This simultaneously opens an
epistemological opportunity to deepen our understanding of how
computation and mathematical manipulation can be carried out by an
intelligent entity \cite{hupkes2019,dziri2023}, and further, how
such reasoning capability is formed, what conditions are required,
and what ``mental apparatus'' an intelligent system must possess in
order to do mathematics.

Following prior work that applied the GASING literacy pedagogy
(\emph{Gampang, Asyik, dan MenyenaNGkan}, i.e., Easy, Fun, and
Enjoyable) to improve model performance in aspects of linguistic
naturalness \cite{situngkir2026a}, this study
extends that paradigm to the domain of numeracy and mathematical
reasoning by applying training based on GASING-\emph{Math}
\footnote{Gasing Academy \url{https://gasingacademy.org/}}, an approach
to learning mathematics introduced by the Indonesian physicist Yohanes Surya that
emphasizes intuitive conceptual understanding, the recognition of numeracy
patterns, and the cultivation of interest in mathematical structure from the
foundational stage onward.
In our implementation, we integrate the GASING method with
\emph{Chain-of-Thought} (CoT) as a technique that has been widely
practiced for facilitating step-by-step reasoning in text-based
generative language models. Here, GASING is applied as the
procedural basis for forming the training and inference patterns
based on CoT when the model is faced with
solving arithmetic problems.

This paper begins by formulating a framework and mathematical
instruments for guiding and analyzing the model's behavior during
training. The initial discussion focuses on the conceptualization of CoT,
the Transformer architecture, and the relationship between the two in
causal language modeling. The following section discusses the
construction of the GASING method as a computational process that
underlies the synthesis of textual representations serving as a
supervisory reference for the model during training. Experiments are then
conducted on the GPT-2 decoder architecture \cite{radford2019} with
TOBA tokenization, which was developed as a syllabic-agglutinative
tokenization scheme for Indonesian and the
Austronesian languages \cite{lumbantobing2026}. This accommodates
the integration of natural language, particularly Indonesian, into the
procedure for solving arithmetic problems.

Throughout the progression of training, we highlight the emergence of
learning phases and elaborate on the development of the model's
arithmetic capability as the number of training steps increases. Through
this, we find that small-scale models can nonetheless learn patterns of
procedural reasoning even through a simple training objective, namely
predicting the next token. Repeated exposure to token continuations that
encode the trajectory of problem solving effectively drives the model
to induce information dependencies, internalize the structure of the
process, and simulate that procedure in solving problems.
In addition, to measure the model's final performance, we also
evaluate the accuracy of solutions to similar arithmetic problems that
were not included in the training data. This evaluation aims to
assess the model's generalization capability rather than mere
reproduction of examples it has already seen. This performance test also
compares the model against various other large language models as an
external benchmark. The results of this study show that training based
on mathematics pedagogy on a Transformer model with a relatively small
number of parameters can yield competitive arithmetic performance, which
can even surpass the performance of models at a larger parameter scale on
the same task domain.

\section{Methods}

\subsection{Chain of Thought}

For a specification \(x \in \mathcal{X}\) within a class
\(\mathcal{X}\), we may consider the existence of a computational process
\(\gamma:\mathcal{X} \rightarrow \mathcal{Y},\gamma \in \Gamma\), namely
a mechanism governed by a set of principles or rules that operates
to produce an output response \(y \in \mathcal{Y}\) for
\(x\). In general, \(\gamma\) can be represented as a state
transition system,
\(\gamma = (\mathcal{S}\ ,\mathcal{A}\ ,T,R)\). The execution of this process
begins from an initial state \(s_{0} \in \mathcal{S}\), which is then
followed by a series of operations or instructions
\({\{ a}_{k}{\}}^{K-1}_{k = 0},a_{k} \in \mathcal{A}\ \) applied
incrementally to the states \({\{ s}_{k}{\}}^{K - 1}_{k = 0}\).
The state update \(s_{k + 1}\) for \(k = 0,...,K - 1\) proceeds
according to the mapping
\(T:\mathcal{S}\  \times \mathcal{A}\  \rightarrow \mathcal{S}\ \)
until reaching the terminal condition at step \(K\), which marks the
culmination of the process execution. The read-out function
\(R:\mathcal{S}\  \rightarrow \mathcal{Y}\) then maps the final state
\(s_{K}\) to produce the semantic output \(y = R(s_{K})\) for
the specification \(x\). Within this framework, we can express the
execution trace of \(\gamma\) for a specification \(x\) as a sequence
of local transitions as follows:

\begin{equation}
\tau_{\gamma}(x) = (s_{k},a_{k},s_{k + 1})_{k = 0}^{K - 1}
\end{equation}

This trace is a procedural object capturing the internal dynamics that may
contain intermediate steps, local computations, or subroutines during
the execution of the process \(\gamma\). If we have a symbolic
vocabulary \(\Sigma\), that is, the set of tokens in a language, the trace
\(\tau_{\gamma}\) can be represented as a textual sequence
through a serialization mapping
\(\sigma:\cup_{K \geq 1}(\mathcal{S} \times \mathcal{A} \times \mathcal{S})^{K} \rightarrow \Sigma^{*}\),
which can then be written as:

\begin{equation}
C(x) = \sigma(\tau_{\gamma}(x))
\end{equation}

In the context of modern text-based generative language modeling,
\(C(x)\) can be viewed as an idealization of the ``\emph{chain of
thought}'' (CoT), namely the textual representation of the intermediate
steps that accompany the computational process \(\gamma\). In other words,
CoT can be seen as a form of symbolic externalization of part of the
procedural structure relevant to the formulation of the final answer.

Language model training generally does not directly ``receive''
the blueprint of the program \(\gamma\), but rather operates through
exposure to token sequences serializing the execution of \(\gamma\). Formally,
in the application of CoT-based training, the textual response to a
specification \(x\) underpinned by a latent program \(\gamma\) can be
written as follows:

\begin{equation}
Z_{\gamma}(x) = C(x)||A(x)
\end{equation}

where \(||\) denotes the concatenation operation between the CoT sequence
\(C(x)\) and the token sequence \(A(x)\) that represents the final
output \(\gamma(x) = y\). In language model training, the sequence of
specification--CoT pairs \((x,Z)\) is often modeled as arising
stochastically by assuming a reference distribution \(P\). This
construction guides supervised training toward finding a generative
model \(Q_{\theta}\) that can produce symbol sequences
with minimal distributional difference from the reference sequences
drawn from \(P\). This distributional difference is generally formulated
through the Kullback--Leibler divergence as follows:

\begin{equation}
D_{KL}(P||Q_{\theta}) = H(P,Q_{\theta}) - H(P)
\end{equation}

where \(H(P,Q_{\theta})\) is the \emph{cross-entropy} between the reference
\(P\) and the model \(Q_{\theta}\), while \(H(P)\) is the entropy of
\(P\), which constitutes the theoretical lower bound of information
compression. Since \(H(P)\) does not depend on the model
parameterization \(\theta\), optimization with respect to \(Q_{\theta}\)
in practice reduces to the minimization of the \emph{cross-entropy}:

\begin{equation}
\theta^{\star} = \arg\min_{\theta}H(P,Q_{\theta}) = \arg\min_{\theta}( - E_{(X,Z)\sim P}\lbrack\log Q_{\theta}(Z|X)\rbrack)
\end{equation}

where \(E_{(X,Z) \sim P\ }\) denotes the expectation over the occurrence
of the sequence \((X,Z)\) from \(P\). This expectation can then be
approximated empirically through the average over sampled sequences during
training. Accordingly, the more commonly used form of the \emph{loss}
is:

\begin{equation}
L(\theta) \approx - \frac{1}{N}\sum_{i = 1}^{N}\log Q_{\theta}(Z^{(i)}|x^{(i)})
\end{equation}

The chosen form of \(Q\) generally takes into account the information
dependencies within the sequence \(Z = (z_{1},...z_{T}) \in \Sigma^{*}\),
where the sequence probability can be factorized causally as:

\begin{equation}
Q_{\theta}(Z|x) = \prod_{t = 1}^{T}q_{\theta}(z_{t}|x,z_{< t})
\end{equation}

where \(z_{< t} = (z_{1},...,z_{t - 1})\) denotes all tokens
in the historical window or context preceding the \(t\)-th token.
In autoregressive inference, the occurrence of each new token
becomes conditionally dependent on the presence of the preceding
tokens. The tokens that have appeared in the context can thereby
influence the distribution of the tokens that will follow. This is what
then gives rise to the ``\emph{scratchpad}
\emph{effect}'' \cite{nye2021}, in which information from a
local or intermediate state or computation that has been
externalized as a token sequence can be reused in the
next reasoning step. If we represent the entire output sequence
as \(Z = C||A\), then its conditional probability
can be factorized as follows:

\begin{equation}
Q_{\theta}(Z|x) = Q_{\theta}(C|x)Q_{\theta}(A|C,x)
\end{equation}

This factorization shows the mathematical relationship between CoT and
the formation of the final answer. The answer tokens can be distributed
conditionally on the information that has been made explicit through the
CoT tokens in the working memory (context window), making CoT an object
that mediates the input specification \(x\) with the production of the
final answer \(A\). Informationally, the presence of this mediating
object can reduce the model's epistemic uncertainty while also lowering
the burden of heuristic search during the inference process. If
\({\overline{H}}_{\theta}\) is the conditional entropy induced
by the model \(Q_{\theta}\), then the following information identity
holds:

\begin{equation}
{\overline{H}}_{\theta}(A|X) - {\overline{H}}_{\theta}(A|X,C) = I_{\theta}(A;C|X) \geq 0
\end{equation}

where the mutual information \(I_{\theta}(A;C|X)\) is non-negative and
takes the value \(0\) only when \(A\) is conditionally independent
of \(C\) given the specification \(X\), under the distribution
induced by the model \(Q_{\theta}\). To the extent that the CoT sequence \(C\) contains
information relevant to the final answer, conditioning on \(C\) will
always narrow the model's predictive distribution in producing
the response \(A\).

Based on this formulation, we can highlight the criteria for a CoT in
text-based language modeling through two principal frameworks, namely
the procedural framework and the informational framework. In the first,
the text sequence in the CoT must be interpretable as the trace of a
latent computational process, at least at a certain level of abstraction.
In other words, it possesses an internal structure that can be aligned
with the computational process assumed to underlie the response through
the preservation of the transition structure or procedural relations among the initial state,
the intermediate operations, and the final state. The final output or
response emerges merely as a consequence of preserving that structure. In the
second framework, which is distributional with respect to the model
\(Q_{\theta}\), the CoT must carry information conditional on the final
response and thereby alter the model's predictive distribution. A CoT is
informationally effective insofar as the tokens within it make the
production of the response more directed and conditioned on the relevant
intermediate information.

\subsection{Information Routing in Language Models}

The representation of the computational process discussed above can also be
constructed in terms of information flow or dependencies. Let
\(V\) be a finite set of sub-processes within the program \(\gamma\)
at a certain level of abstraction, and let \(X_{v}\) be the
computational variable at \(v \in V\). We can define a directed
acyclic graph (DAG) \(G_{\gamma} = (V,E)\) with
\(E = \{(u,v),\ u,v \in V\}\) being the set of directed relations between two
sub-processes whenever \(X_{v} = f_{v}({...,X}_{u},...)\), that is, \(X_{u}\)
is a direct argument required to compute \(X_{v}\).
This graph is a representation of the functional information route of the
program \(\gamma\)'s execution, with each relation within it indicating
information dependency among local or
inter-stage blocks of computation; that is, which variable must be passed on to
compute another variable at the next stage.

If \(C = \sigma(\tau_{\gamma}(x)) = (c_{1},..c_{T})\) is the
textual serialization of the execution trace of the program \(\gamma\)
interpreted as the CoT for a specification \(x\), then there exists
a partial alignment map
\(\phi:V \rightharpoonup \{(s,e):1 \leq s \leq e \leq T\}\) that
connects a \emph{span} in the text \(C\) with a sub-process node
on the graph \(G_{\gamma}\). With this mapping, a graph
\({G^{\text{span}}_{\gamma}} = (\phi(V),\ E^{\text{span}})\) can be defined to
represent the information route on the textual surface, the CoT \(C\),
which satisfies:

\begin{equation}
(\phi(u),\phi(v)) \in E^{\text{span}} \Rightarrow (u,v) \in E
\end{equation}

This graph becomes important for the purpose of analyzing the behavior of
language models based on the Transformer architecture \cite{vaswani2017},
given that the \emph{self-attention} mechanism in
Transformers operates on token representations and their positional
indices in the sequence. For a layer \(l\) of the model, the representation of
token \(c_{i}\) is expressed as a hidden-state vector
\({h_{c_{i}}^{(l)}} \in \mathbb{R}^{d}\). \emph{Self-attention} forms
inter-token interactions through a linear transformation of the \emph{query}
(Q), \emph{key} (K), and \emph{value} (V):

\begin{equation}
Q^{(l)} = H^{(l)}{W_{Q}^{(l)}}, K^{(l)} = H^{(l)}{W_{K}^{(l)}}, V^{(l)} = H^{(l)}{W_{V}^{(l)}}
\end{equation}

where \(H = \lbrack{h_{1}^{(l)}},...,{{h_{T}^{(l)}}\rbrack}^{\top}\) and
\({W_{Q}^{(l)}}\), \({W_{K}^{(l)}}\), \({W_{V}^{(l)}}\) are the trained
parameters at that layer. The magnitude of the attention of a token \(c_{i}\)
toward a token \(c_{j}\) is then given by the equation:

\begin{equation}
{\alpha^{(l)}_{c_{i},c_{j}}} = \text{softmax}(\frac{{{q^{(l)}_{c_{i}}}}{(k_{c_{j}}^{(l)}})^{\top}}{\sqrt{d_{k}}})
\end{equation}

where \(d_{k}\) is the dimension of the \emph{key/query} vector, used
as a normalization factor to maintain the stability of the gradient scale.
In particular, in autoregressive decoder architectures such as GPT, a token
\(c_{i}\) is restricted to attend only to tokens \(c_{j}\) at positions
that precede it or to itself (\(j \leq i\)) through
\emph{causal masking}. The representation of \(c_{i}\) at the next layer
\(l + 1\) is then computed as a weighted combination over all
tokens preceding \(c_{i}\):

\begin{equation}
{h_{c_{i}}^{(l + 1)}} = \sum_{j = 1}^{i}{\alpha^{(l)}_{c_{i},c_{j}}}{v^{(l)}_{j}}
\end{equation}

Through inference based on the Transformer architecture, each relation
\(((\phi(u),\phi(v)) \in E^{\text{span}}\) on the graph \({G^{\text{span}}_{\gamma}}\)
can be assigned a weight value corresponding to the magnitude of
aggregate \emph{attention} between \emph{span} nodes. Empirically, in the
model \(Q_{\theta}\), the weight function
\({w^{(l)}_{\theta}}:E \rightarrow \mathbb{R}\) for a layer \(l\) is computed
according to the following formulation:

\begin{equation}
w^{(l)}_{\theta}(\phi(u),\phi(v)) = \frac{1}{|\phi(u)||\phi(v)|}\sum_{i \in \phi(v)}\sum_{j \in \phi(u)}{\alpha^{(l)}_{c_{i},c_{j}}}
\end{equation}

where \(|\phi(u)|\) and \(|\phi(v)|\) denote the respective lengths of the
\emph{spans}. This formulation provides not only an observational
instrument but can also serve as an object of measurable intervention for
observing and interpreting the model's behavior mechanistically.
Using the graph \({G^{\text{span}}_{\gamma}}\) as an ideal reference of program
execution, controlled interventions can be performed on a trained
Transformer model to test whether the CoT generated through inference
satisfies procedural alignment and informational effectiveness, or
is merely a textual decoration. One such analysis that can be
performed is to apply suppression control or \emph{blocking} on the
\emph{attention} mechanism using the \emph{attention masking} technique
\cite{bogdan2025,saha2025}. If
\((\phi(u),\phi(v)) \in E^{\text{span}}\), in other words the tokens
on \(\phi(v)\) ideally depend informationally on the text \emph{span}
\(\phi(u)\), then we can test the empirical question of whether
the presence of the text on \(\phi(u)\) effectively reduces the model's
predictive uncertainty more than another, irrelevant \emph{span}
\(\phi(u')\), \((\phi(u'),\phi(v)) \notin E^{\text{span}}\),
when the model generates the text on \(\phi(v)\).

Formally, we can define an intervention operator
\(M^{(u,v)} \in \{ 0,1\}^{T \times T}\), with \({M^{(u,v)}}_{ij} = 1\)
if and only if \(j \in \phi(u)\) and \(i \in \phi(v)\), and
\(0\) for the other elements. For each layer \(l\), applying
\(M^{(u,v)}\) modifies the \emph{attention} score matrix before
the \emph{softmax} operation as follows:

\begin{equation}
s^{(l)}_{c_i, c_j} = 
\begin{cases} 
-\infty, & \text{if } M^{(u,v)}_{ij} = 1 \text{ or } j > i \\
\frac{q^{(l)}_{c_i} (k_{c_j}^{(l)})^\top}{\sqrt{d_{k}}}, & \text{otherwise}
\end{cases}
\end{equation}

Replacing the pre-\emph{softmax} score with \(- \infty\) makes the
\emph{attention} value \({\alpha^{(l)}_{c_{i},c_{j}}} = 0\), so that
all information from the tokens within the span \(\phi(u)\) is
effectively cut off from the computation of the representation
\({h_{c_{i}}^{(l + 1)}},i \in \phi(v)\). In the construction of the dependency
graph \({G^{\text{span}}_{\gamma}}\), this algebraic manipulation is equivalent
to setting the relation weight
\({w^{(l)}_{\theta}}(\phi(u),\phi(v)) = 0\) at every layer \(l\) in the
model \(Q_{\theta}\). Because the intervention is applied directly to the
\emph{attention} distribution, the network structure and trained parameters
of the model do not change; the modification occurs only in the internal
information flow for a given inference.

The effect of the intervention \(M^{(u,v)}\) on the model's generative
process can then be measured through the change in \emph{loss} on the
target tokens within the span \(\phi(v)\), formulated as the difference in
\emph{negative log-likelihood} (NLL) relative to the \emph{baseline} state
as follows:

\begin{equation}
\Delta_{\phi(u) \rightarrow \phi(v)} = ( - \sum_{i \in \phi(v)}\log q_{\theta}(c_{i}|x,c_{< i};M^{(u,v)}) - ( - \sum_{i \in \phi(v)}\log q_{\theta}(c_{i}|x,c_{< i}))
\end{equation}

If a procedural relation underpinned by a program \(\gamma\)
is genuinely internalized by the model, then that relation should
be manifested observationally through the \emph{attention} pattern and
predictive sensitivity to the \emph{masking} intervention \(M^{(u,v)}\)
according to the textual-surface graph \({G^{\text{span}}_{\gamma}}\). The magnitude of
\(\Delta_{\phi(u) \rightarrow \phi(v)}\) can then serve as an operational proxy
for showing the extent to which the presence of information on \(\phi(u)\)
contributes to narrowing the model's predictive distribution over the
tokens in \(\phi(v)\). Furthermore, blocking the \emph{attention}
from the tokens within \(\phi(v)\) toward \(\phi(u)\) idealized
in the relation \(((\phi(u),\phi(v)) \in E^{\text{span}}\) should produce
a greater predictive degradation than suppression on
another, irrelevant \emph{span}. To quantify this difference in
response, the information contrast can be formulated as follows:

\begin{equation}
\Gamma(v;u,u') = \Delta_{\phi(u) \rightarrow \phi(v)} - \Delta_{\phi(u') \rightarrow \phi(v)}
\end{equation}

where \(\phi(u')\) is a node on the graph \({G^{\text{span}}_{\gamma}}\)
without a direct relation to \(\phi(v)\),
\((\phi(u'),\phi(v)) \notin E^{\text{span}}\), but having a positional range
that precedes \(\phi(v)\). A positive value of \(\Gamma(v;u,u')\)
indicates that the model has formed a dependency or route of
information propagation through the CoT that not only plays a causal role in
forming the final answer, but is also aligned with the procedural
structure of the program \(\gamma\).

\section{The GASING Method for Basic Arithmetic}

Unlike conventional approaches to learning computation, the GASING method
solves basic arithmetic problems by applying a ``left-to-right'' process,
in which the decomposition of the computation begins from the position of the largest digit
to the smallest. This distinctive feature is not merely a pedagogical
variation but also carries computational consequences in the practice of language modeling.
In particular, for Transformer decoder-based language models, prior
work \cite{lee2023} has shown that training a model
on addition in the standard format
`\(a_{3}a_{2}a_{1} + b_{3}b_{2}b_{1} = c_{3}c_{2}c_{1}\)' is
suboptimal because it is not aligned with the token inference mechanism
that proceeds causally. The first token of the answer that must
be predicted is \(c_{3}\), the digit at the largest position of
the result. Yet the value \(c_{3}\) can be influenced by the carry chain
arising from the digits at smaller positions. In other words, the model
is required to produce the digit that appears earliest textually,
yet that can only be determined algorithmically after the information from
the entire right-hand side has been taken into account. One way this
problem is addressed is by reversing the order of the result digits,
`\(a_{3}a_{2}a_{1} + b_{3}b_{2}b_{1} = \$ c_{1}c_{2}c_{3}\$\)',
accompanied by symbols marking the beginning and end of the reversed
result. In computation with the GASING approach, such a modification is not performed because
its procedure is already aligned with the causal order of token generation.

Let us fix a number base \(B\); we can then represent the digits of
a non-negative number \(A = \sum_{i = 1}^{n}a_{i}B^{n - i}\) as a sequence from left to
right, \(\text{dig}_{n}(A) = (a_{1},...a_{n}),\ a_{i} \in \{ 0,...,B - 1\}\),
with the addition of zero \emph{padding} on the left side if the digit length
needs to be equalized. While a computation is in progress, the GASING approach maintains
a \emph{state} \(S = (s_{1},...,s_{m})\) as a temporary representation
of the answer being built up to step \(m\). When the
computation on a digit at the smaller position at \(m + 1\)
yields an evaluated value \(x\), that digit is appended to the right
side of the sequence:

\begin{equation}
S \oplus x = (s_{1},...,s_{m},x)
\end{equation}

The difference between GASING and the conventional approach lies in the treatment
of values that exceed the base (\(x \geq B\) in the cases of addition
and multiplication) or that are in deficit (\(x < 0\) in subtraction).
Normalization to handle the carry (\emph{carry}) and the borrow (\emph{borrow})
is applied as a correction operation on the digits at the
larger positions that were previously formed as elements of the sequence
\(S\). When the evaluation at position \(j\) yields a \emph{carry}
\(x \geq B\), the correction is performed on the preceding element of the
\emph{state} \(S\) as follows:

\begin{equation}
\begin{aligned}
s_{j - 1} &\leftarrow s_{j - 1} + \lfloor s_{j}/B \rfloor \\
s_{j} &\leftarrow s_{j} \bmod B
\end{aligned}
\end{equation}

Meanwhile, in the case of subtraction, when \(x < 0\) or a value
deficit occurs at position \(j\), the borrow correction operation (\emph{borrow})
is applied to the digit at the larger position. But because the
digit to the left at position \(j - 1\) has already been computed, the
correction operation can be performed retrospectively on the \emph{state}
as follows:

\begin{equation}
\begin{aligned}
s_{j - 1} &\leftarrow s_{j - 1} - 1 \\
s_{j} &\leftarrow s_{j} + B
\end{aligned}
\end{equation}

To guide language model training according to the GASING approach, we
need to implement it as a program that can be executed in a
computational environment. The execution of this program is what then
produces the computational trace that can be serialized textually
in natural-language articulation. With this operationalization, GASING
becomes the procedural basis that determines how an arithmetic
problem is decomposed, while the trace of the process's execution is the CoT that
serves as the supervisory medium encoding the trajectory of that
decomposition so that it can be learned by a generative language model. This CoT
is then formatted to mediate the question--answer pair as a
demonstration of working or an externalization of step-by-step thinking.

\subsection{Addition and Subtraction}

The addition operation with the GASING method is carried out by first
equalizing the digit lengths of the two \emph{operands} \(A\) and \(C\)
into \((a_{1},...,a_{n})\) and \((c_{1},...,c_{n})\). The iteration
then proceeds from \(i = 1\) to \(n\), where for each
position \(i\), the local computation performed is
\(m = a_{i} + c_{i}\). If \(m < B\), the value is directly
appended to the right end of the sequence representing the temporary
result \(S\). Meanwhile, when \(m \geq B\), the value \(m\) is split into two
components, namely \(h = \lfloor m/B\rfloor\) and
\(k = m \bmod B\). The component \(h\) is then added to the
last element of the existing sequence, while \(k\) is placed
as a new element on the right side of the sequence. At the end of an iteration,
the \emph{carry} correction operation is applied to repair possible
internal \emph{overflow} in the preceding elements within the sequence.

\begin{algorithm}[t]
\small
\caption{GASINGSequenceAddition}
\begin{algorithmic}[1]
\Require Non-negative integers $A, C$, base $B=10$
\Ensure Sum $Y=A+C$
\State $n \gets \max(\lambda_B(A), \lambda_B(C))$
\State $(a_1, \ldots, a_n) \gets \operatorname{digits}_B(A, n)$
\State $(c_1, \ldots, c_n) \gets \operatorname{digits}_B(C, n)$
\State $S \gets ()$
\For{$i=1, \ldots, n$}
    \State $m \gets a_i + c_i$
    \If{$S=()$}
        \State $S \gets (m)$
    \ElsIf{$m < B$}
        \State $S \gets \operatorname{append}(S, m)$
    \Else
        \State $h \gets \lfloor m/B \rfloor$
        \State $k \gets m \bmod B$
        \State Let $\ell \gets |S|$
        \State $s_\ell \gets s_\ell + h$
        \State $S \gets \operatorname{append}(S, k)$
        \State $S \gets \Call{CarryNormalize}{S, B}$
    \EndIf
\EndFor
\State $Y \gets \operatorname{concat}_B(S)$
\State \Return $Y$
\end{algorithmic}
\end{algorithm}

GASING applies a similar procedure for the subtraction operation, but
using a complement mechanism. If at position \(i\)
\(a_{i} \geq c_{i}\) holds, then the new element is directly
written as \(r = a_{i} - c_{i}\). But if
\(a_{i} < c_{i}\), the subtrahend digit \(c_{i}\)
is complemented with respect to the base by \(p = B - c_{i}\), then the
new element is computed as \(r = a_{i} + p\), and at the same time
the last element of the existing sequence is decreased by one. If
this decrement by one makes the internal element negative,
the \emph{borrow} correction operation is applied to repair it by
pushing the \emph{borrow} further to the left retrospectively.

\begin{algorithm}[t]
\small
\caption{GASINGSequenceSubtraction}
\begin{algorithmic}[1]
\Require Integers $A > C \geq 0$, base $B=10$
\Ensure Difference $Y = A - C$
\State $n \gets \max(\lambda_B(A), \lambda_B(C))$
\State $(a_1, \ldots, a_n) \gets \operatorname{digits}_B(A, n)$
\State $(c_1, \ldots, c_n) \gets \operatorname{digits}_B(C, n)$
\State $S \gets ()$
\For{$i=1, \ldots, n$}
    \If{$a_i \geq c_i$}
        \State $r \gets a_i - c_i$
        \State $S \gets \operatorname{append}(S, r)$
    \Else
        \State \textbf{assert} $S \neq ()$
        \State $p \gets B - c_i$
        \State $r \gets a_i + p$
        \State Let $\ell \gets |S|$
        \State $s_\ell \gets s_\ell - 1$
        \State $S \gets \operatorname{append}(S, r)$
        \State $S \gets \Call{BorrowNormalize}{S, B}$
    \EndIf
\EndFor
\State $S \gets \operatorname{trim}_0(S)$
\State $Y \gets \operatorname{concat}_B(S)$
\State \Return $Y$
\end{algorithmic}
\end{algorithm}

\subsection{Multiplication}

The multiplication operation with the GASING method is not performed by
accumulating partial results from the right side as in the
conventional approach. For \emph{operands} \(A\) and \(C\) whose
digit representations have lengths \(n\) and \(m\) respectively, each pair
\((a_{i},c_{j})\) is multiplied to produce
\(t_{ij} = a_{i}c_{j}\). The result of this computation is placed
in the position group \(g_{ij} = (n - i) + (m - j)\), which expresses the
base power of that digit product. All evaluated values in the same position
group are then summed as follows:

\begin{equation}
M_{g} = \sum_{(i,j):g_{ij} = g}t_{ij}
\end{equation}

\begin{algorithm}[t]
\small
\caption{GASINGGroupSequenceMultiplication}
\begin{algorithmic}[1]
\Require Non-negative integers $A, C$, base $B=10$
\Ensure Product $Y = A \cdot C$
\State $(a_1, \ldots, a_n) \gets \operatorname{digits}_B(A, \lambda_B(A))$
\State $(c_1, \ldots, c_m) \gets \operatorname{digits}_B(C, \lambda_B(C))$
\ForAll{digit pairs $(i, j)$}
    \State $t_{ij} \gets a_i c_j$
    \State $g_{ij} \gets (n-i) + (m-j)$
\EndFor
\ForAll{place powers $g$ appearing among the $g_{ij}$}
    \State $M_g \gets \sum_{(i,j): g_{ij}=g} t_{ij}$
\EndFor
\State Order the place powers as $g_1 > g_2 > \cdots > g_K$
\State $S \gets (M_{g_1})$
\For{$\ell = 2, \ldots, K$}
    \State $u \gets M_{g_\ell}$
    \State $h \gets \lfloor u/B \rfloor$
    \State $k \gets u \bmod B$
    \State Let $\ell_S \gets |S|$
    \State $s_{\ell_S} \gets s_{\ell_S} + h$
    \State $S \gets \operatorname{append}(S, k)$
    \State $S \gets \Call{CarryNormalize}{S, B}$
\EndFor
\State $Y \gets \operatorname{concat}_B(S)$
\State \Return $Y$
\end{algorithmic}
\end{algorithm}

The answer representation \(S\) is built iteratively by processing
the group values \(M_{g}\) in order from the largest power to the
smallest (\(g_{1} > g_{2} > ... > g_{K}\)). This process is
initiated by setting the group value at the largest position,
\(M_{g_{1}}\), as the first element in the
sequence \(S\). For each subsequent group
\(M_{g_{l}}\) (with \(l = 2,..,K\)),
as is also done in the addition operation, the value \(M_{g_{l}}\)
is split into a \emph{carry} \(h = \lfloor M_{g_{l}}/B\rfloor\) and a remaining
digit \(k = M_{g_{l}} \bmod B\). The component \(h\) is then
added to the last element of the already-formed sequence \(S\),
and \(k\) becomes a new element at the right end of the sequence.
The \emph{carry} correction operation is then applied at the end of each
iteration to evaluate the sequence retrospectively toward the left.

\subsection{Division}

Division has a slightly different structure because it involves a
temporary target and a remainder, but it still maintains a procedural direction
consistent with the left-to-right principle. To divide a number
\(A\) by \(C\) satisfying \(A \geq C\), the digits of the number \(A\)
are first read from the leading position until they form an initial target
\(T \geq C\) with digit representation \((a_{1},...,a_{j})\). At
each iteration step, the quotient digit \(q\) is determined through
estimation and gradual decrement until a value is obtained that satisfies
\(qC \leq T\). The product \(qC\) is then subtracted from the target to
produce the remainder (\emph{remainder}) \(R = T - qC\). If there
remains an unprocessed digit \(a_{j + 1}\), that digit is
sequentially taken to form a new computation target
\(T \leftarrow BR + a_{j + 1}\). Each digit \(q\) is then appended
to the sequence representing the temporary answer \(S\) in order.
In the case where \(T < C\), the evaluated value \(q\) is directly
zero.

\begin{algorithm}[t]
\small
\caption{GASINGFrontDigitDivision}
\begin{algorithmic}[1]
\Require Integers $A \geq C > 0$, base $B=10$
\Ensure Quotient $Q$ and remainder $R$ such that $A = CQ + R$
\State $(a_1, \ldots, a_n) \gets \operatorname{digits}_B(A, \lambda_B(A))$
\State $h \gets \min\{j : \operatorname{value}(a_1, \ldots, a_j) \geq C\}$
\State $T \gets \operatorname{value}(a_1, \ldots, a_h)$
\State $q_{\mathrm{seq}} \gets ()$
\State $\mathit{cursor} \gets h$
\While{\textbf{true}}
    \If{$T < C$}
        \State $q \gets 0$
        \State $P \gets 0$
        \State $R \gets T$
    \Else
        \State $f \gets \Call{FrontPart}{T, C, B}$
        \State $\beta \gets$ leading digit of $C$
        \State $e \gets \min(B-1, \max(1, \lfloor f/\beta \rfloor))$
        \State $q \gets e$
        \While{$qC > T$}
            \State $q \gets q - 1$
        \EndWhile
        \State $P \gets qC$
        \State $R \gets T - P$
    \EndIf
    \State $q_{\mathrm{seq}} \gets \operatorname{append}(q_{\mathrm{seq}}, q)$
    \If{$\mathit{cursor} = n$}
        \State \textbf{break}
    \EndIf
    \State $\mathit{cursor} \gets \mathit{cursor} + 1$
    \State $T \gets BR + a_{\mathit{cursor}}$
\EndWhile
\State $Q \gets \operatorname{concat}_B(q_{\mathrm{seq}})$
\State \Return $(Q, R)$
\end{algorithmic}
\end{algorithm}

\section{Language Model Training}

Based on the implementation of GASING as a computational procedure,
we build a textual dataset that serves as the source of supervision for
running the language model training process. This dataset contains samples in an
instructional format covering the solution of basic arithmetic problems on base-10
integers up to a maximum length of three digits. The arithmetic problems
are presented in the form of natural-language questions to encourage the model
to learn the mapping of linguistic expressions to the appropriate computational
procedure. Before the final answer is given, the target output includes
\emph{Chain-of-Thought} (CoT) reasoning based on the
execution trace of the GASING computational procedure and articulated through the syntax
and lexicon of natural language. To familiarize the model with the
variety of surface realizations of semantically identical numeric concepts,
the dataset also combines the use of numeric symbols and of
numbers written in word form. As an example,
an \emph{operand} can be written either in numeric form (e.g.,
`\emph{123}') or in the corresponding word form in Indonesian
(e.g., `\emph{seratus dua puluh tiga}').

The dataset used consists of 90,000 training examples with a
balanced distribution across the four arithmetic operations. In addition,
we ensure the uniqueness of each sample by making sure that no
operation with the same \emph{operand} pair appears more than once.
For the commutative operations, namely addition and multiplication, the
\emph{operand} pairs are converted to a canonical form to prevent duplication
arising from order permutations. For the subtraction and division operations, the
first \emph{operand} is constrained to always be greater than or equal to the second
\emph{operand} so that only non-negative results and integer
forms with a remainder are involved. Over the entire problem
space, the four types of operations with these constraints provide a total
of 2,001,000 unique arithmetic problems. The training data used
therefore contains only \(< 5\%\) of the entire sample space
that could possibly be explored.

The experiments were conducted using the GPT-2 decoder architecture with 86 million
parameters initialized with random weights. The model uses the TOBA
tokenization scheme with a vocabulary size of 284 tokens covering
Indonesian syllable units, numeric symbols, arithmetic notation,
as well as a number of special tokens used for the purposes of formatting
and controlling output structure. The model architecture
configuration is summarized in Table \ref{tab:table1}.

\begin{table}[h]
\centering
\caption{Model Architecture Specification}
\label{tab:table1}
\begin{tabular}{ccccc}
\toprule
\textbf{Layers} & \textbf{Hidden Size} & \textbf{Attention (Heads / Dim)} & \textbf{MLP (Hidden / Exp)} & \textbf{Max Sequence} \\
\midrule
12 & 768 & 12 / 64 & 3072 / $4\times$ & 1280 \\
\bottomrule
\end{tabular}
\end{table}

Model training was performed with the standard autoregressive language
modeling objective. For a sequence of tokens \(x_{1},x_{2},...,x_{T}\), the model
maximizes the probability of the occurrence of the next token based on the
window of preceding tokens, with the \emph{loss} function:

\begin{equation}
\mathcal{L}\  = - \log\sum_{t = 1}^{T}P(x_{t}|x_{< t})
\end{equation}

where \(P(x_{t}|x_{< t})\) denotes the probability assigned by the model
to the target token at the \(t\)-th position. The training procedure does not
involve the application of \emph{reinforcement learning} or
\emph{reward}-based optimization. In other words, the model is not directly
optimized to obtain mathematical correctness. The optimization objective
is merely to predict the next token over the reasoning trace and
the final answer. To observe the development of capability during the
training process, model \emph{checkpoints} are evaluated periodically
using a test dataset composed of 10,000 samples of similar arithmetic
problems that are exclusive from the training data. The final answer
for a sample is then extracted from the output produced by
the model's inference. The accuracy performance is then computed by comparing
the extracted answer \(\widehat{y}\) against the corresponding \emph{ground
truth} value:

\begin{equation}
\text{Accuracy} = \frac{1}{N}\sum_{i}^{N}\mathbf{1}({\widehat{y}}_{i} = y_{i})
\end{equation}

where \(N\) denotes the number of evaluation samples and \(\mathbf{1}( \cdot )\)
is the indicator function that returns the value 1 if the prediction
is identical to the \emph{ground truth}, and 0 otherwise.

\section{Results}

\subsection{Learning Progression}

\begin{figure}[h]
\centering
\includegraphics[width=13cm]{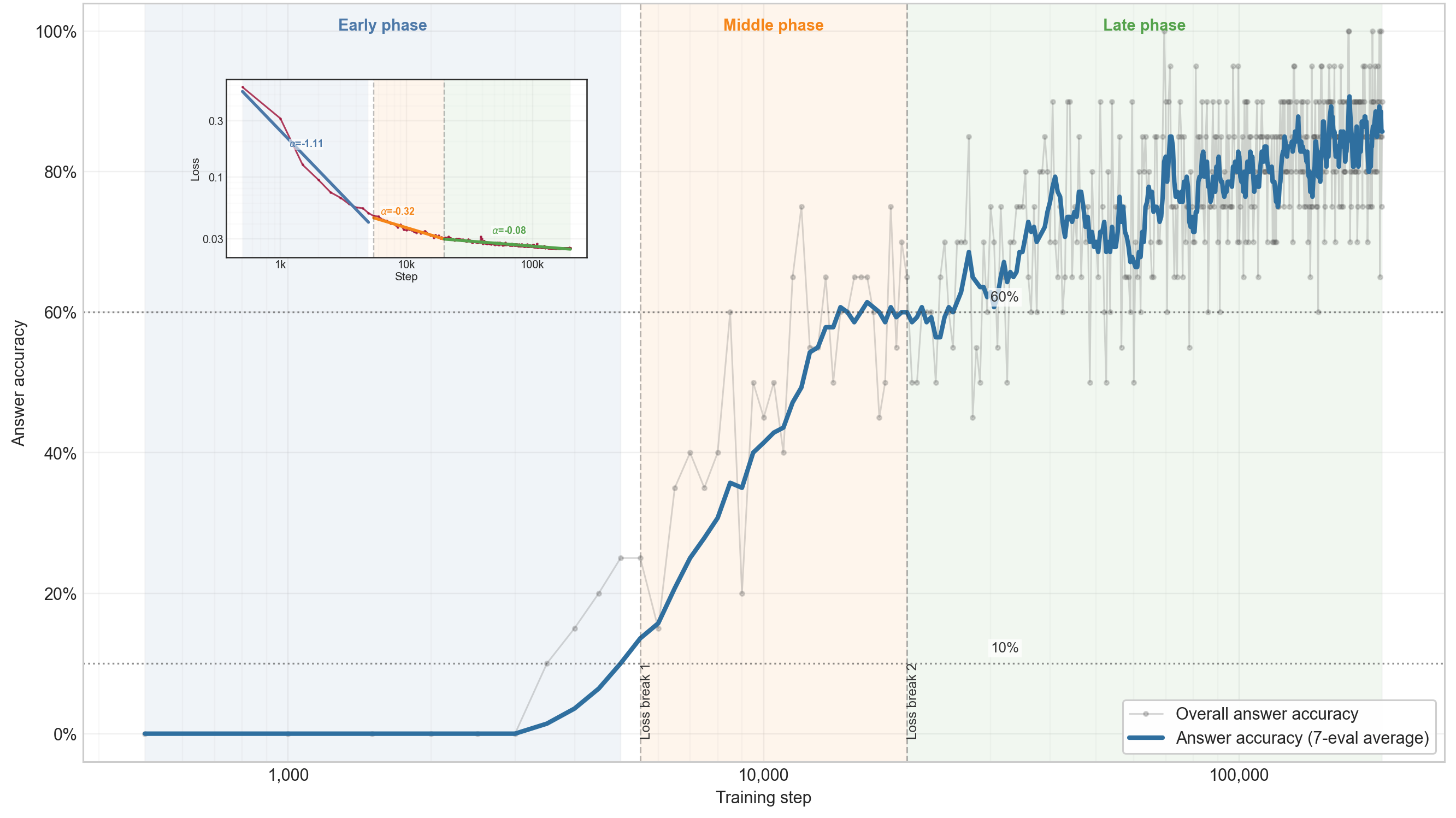}
\caption{Development of the model's arithmetic computation accuracy performance during training.
\textbf{Inset}: Plot of the \emph{cross-entropy loss} evaluation metric during training for the \emph{next-token prediction} (NTP) objective.}
\label{fig:figure1}
\end{figure}

Periodic monitoring of the \emph{cross-entropy loss} on the evaluation data
during training shows the presence of phases in the trajectory
of \emph{loss} change characterized by differences in the exponent of the
\emph{loss}-function \emph{fit}. In the early stage of training, the
\emph{loss} value exhibits a steeply declining trend. However,
as the number of training steps increases, as shown in the \emph{inset} of
Figure \ref{fig:figure1}, the model enters a phase marked by the flattening of the rate of
\emph{loss} decline, reflected in the increasingly larger exponent
\(\alpha\). The steep decline of the \emph{loss} in the early phase can be
interpreted as the period during which the model learns the syntax and
structure of language. The pressure to predict the majority of tokens in
the training data drives the model to recognize linguistic regularity and
acquire the basic capability to ``use language''. This learning phase is
then followed by a middle phase in which the slowing of the \emph{loss}
decline indicates a learning process beginning to focus on
a small number of tokens that constitute critical information. The renewed
slowing of \emph{loss} change in the final phase indicates a process
of maturation of the representations that have been learned. Consistent with
this interpretation, the evaluation of arithmetic problem-solving
capability likewise exhibits three developmental phases that correspond to
the \emph{loss} trajectory. A significant improvement in performance occurs
coinciding with the middle phase, namely when the model's accuracy increases
from a value of less than 10\% to reaching about 60\%.

\begin{figure}[h]
\centering
\includegraphics[width=11.5cm]{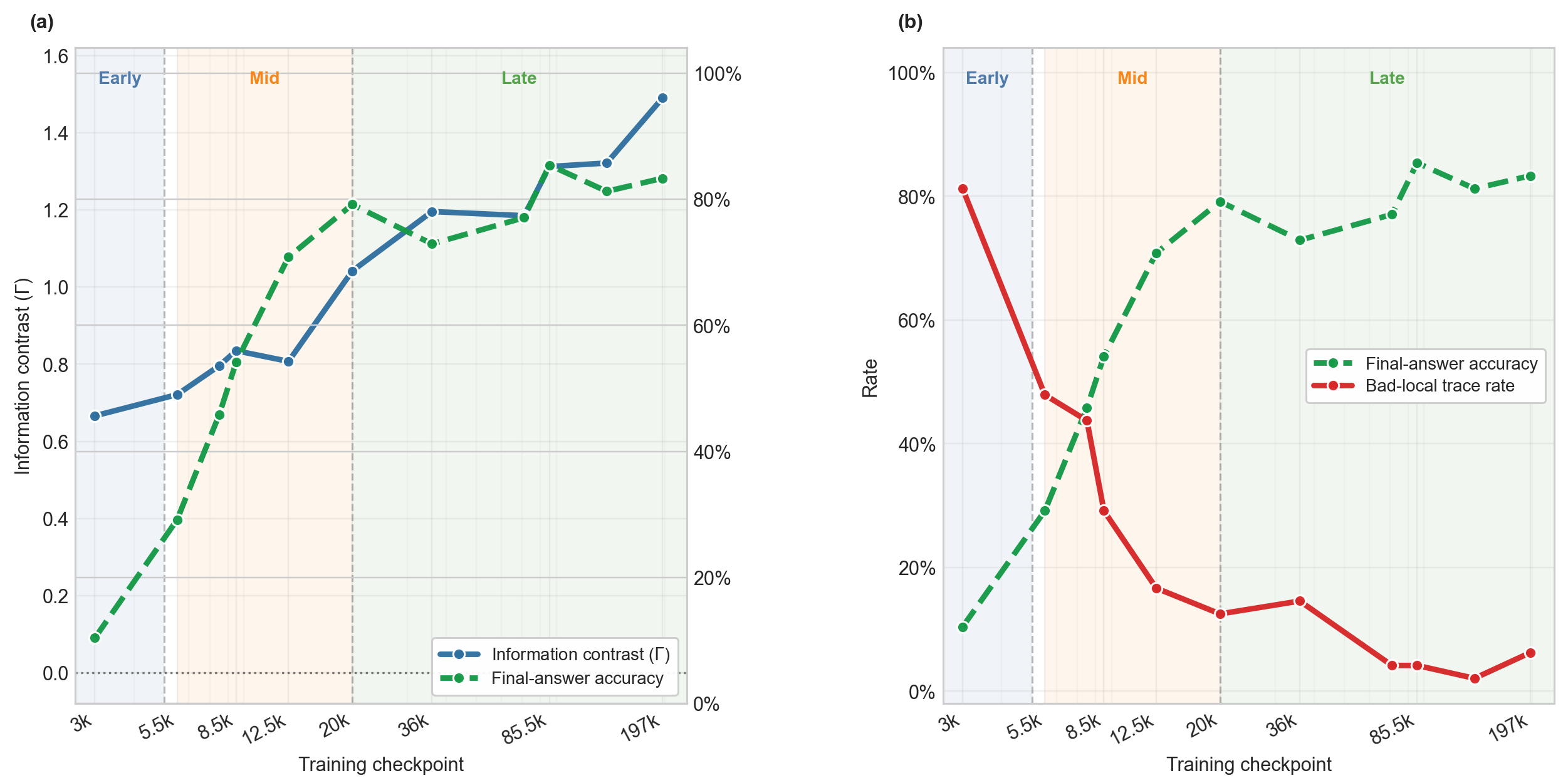}
\caption{(a) Plot of the information contrast value \(\Gamma\) (blue) for each analyzed training \emph{checkpoint}.
This value is obtained by performing \emph{attention} suppression on the CoT sequence produced by the model's inference
for the four types of arithmetic problems tested. (b) Plot of the percentage of occurrence of computation errors (red) in the
CoT produced by inference.}
\label{fig:figure2}
\end{figure}

This finding prompts further investigation into the factors that
cause the surge in accuracy during the middle phase. For this purpose,
we leverage the model weights at a number of training \emph{checkpoints}
and construct a representation of the information route based on the textual
\emph{Chain-of-Thought} (CoT) externalized during inference
before the model produces the final answer. With this, we obtain
a functional information graph that depicts the model's reasoning
structure for each analyzed \emph{checkpoint}. Figure \ref{fig:figure2}(a) shows
that before accuracy increases significantly, the model has first
internalized a procedural pathway consistent with the GASING
method. This is evident from the information contrast that is already
positive, indicating that blocking the computation block that is
relevant for the propagation of information to the next computation stage
results in a greater predictive degradation than
blocking a random pathway in the CoT. Nonetheless, observation of the
functional information graph at this stage identifies that there are still many
local computations that yield incorrect intermediate values (Figure \ref{fig:figure2}(b)).
Specifically, these local computation errors are largely
arithmetic operations on small-digit numbers required for the
solution of longer and more complex problems. These errors in the
intermediate results are then propagated through the reasoning pathway
that has been formed, thereby producing an incorrect final answer
even though it is already correct in terms of its procedural structure.

\begin{figure}[h]
\centering
\includegraphics[width=11.5cm]{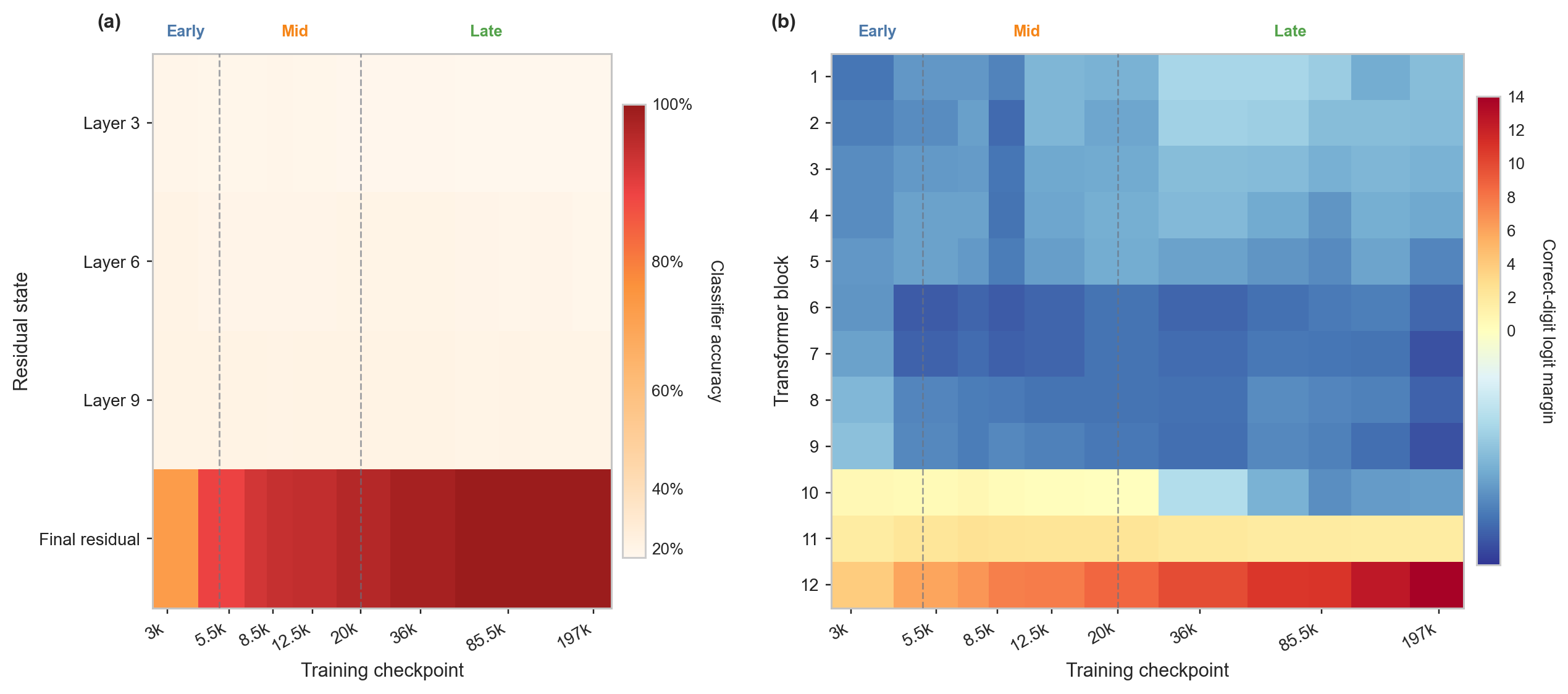}
\caption{(a) \emph{Heatmap} of the prediction accuracy of the correct digit by the \emph{classifier} based on the model's \emph{residual stream} at layers 3, 6, 9, and 12
for the analyzed training \emph{checkpoints}. High values indicate that the internal representation of the correct digit becomes increasingly separated from the other tokens
as a candidate output of the model's inference. (b) \emph{Heatmap} of the \emph{logit} margin of the correct digit relative to the competing digit token based on the \emph{residual stream}
at each \emph{layer} and \emph{checkpoint}.}
\label{fig:figure3}
\end{figure}

During the middle phase, the proportion of local computations that yield
incorrect values continues to decrease, proceeding in tandem with the increase
in final-answer accuracy. To trace the underlying mechanism,
we test the \emph{hidden state} representation of the model
exactly when the model is about to infer the result-digit token of a local computation.
This testing is performed through two approaches. First, we train
a multinomial logistic regression \emph{classifier} to predict the
digit of the local computation result from the \emph{residual stream} at various
layers of the Transformer blocks. The results in Figure \ref{fig:figure3}(a) show that
as training proceeds, the \emph{residual stream} increasingly carries
a clear signal about the result digit, marked by the increase in the
\emph{classifier} accuracy during the middle phase (steps 3K $\to$ 12.5K).
This strengthening is observed to occur primarily at the final-layer blocks. Second, the testing
is performed by applying the \emph{logit-lens} technique \cite{wang2025},
in which the \emph{residual} at each block is passed
through the model's output \emph{readout} to measure the \emph{logit} margin
between the correct digit and the competing digit. As
training proceeds, the \emph{residual stream} at the final-layer blocks
increasingly gives a positive margin to the correct digit (Figure \ref{fig:figure3}(b)).
In other words, the model's internal representation becomes increasingly
selective in surfacing the relevant value as the dominant candidate output
token. The results of this testing show that the main transition
in the middle phase is the improvement of the internal representation that
makes the correct local computation result more available and
more preferred by the model. This phenomenon bears a resemblance to the
``mental arithmetic'' (\emph{mencongak}) process in humans. When an arithmetic pattern occurs
frequently enough and exerts strong predictive pressure, obtaining its result
does not always require an explicit procedure or step-by-step reasoning;
part of the result value can be activated more directly
through the internal representation patterns that have been formed during the learning
process.

Taken together, these findings clarify the picture of the development of
the language model's capability during training. In the early stage, the model
mainly learns the statistical regularities of language---including syntax,
phrase structure, and semantic relations---which serve as the foundation for
the formation of a mental framework for manipulating information.
This mastery of language then enables information to be organized,
maintained in working memory, and manipulated in a structured manner.
As in humans, solving complex problems requires a
representational medium of this kind to support coherent and
organized thinking. On the other hand, solving simpler
problems often no longer requires an explicit procedure,
but instead relies on associative memory acquired through repeated
exposure to data \cite{situngkir2026b}. The closest analogy
is the capacity for mental arithmetic, where the result of a simple operation
can appear directly without consciously executing the steps of
the computation. The development of these two capabilities in the language
model together facilitates more complex arithmetic reasoning. Furthermore,
the results of this training show that the internalization of a procedure does
not always require an algorithmic blueprint that is made explicit. The demand
to accurately predict the continuation of a token sequence can in fact
drive the organization of neuronal activations in the language model to form an
information-dependency structure that functionally resembles a computation
procedure.

\subsection{Evaluation of Arithmetic Capability}

\begin{figure}[h]
\centering
\includegraphics[width=12.5cm]{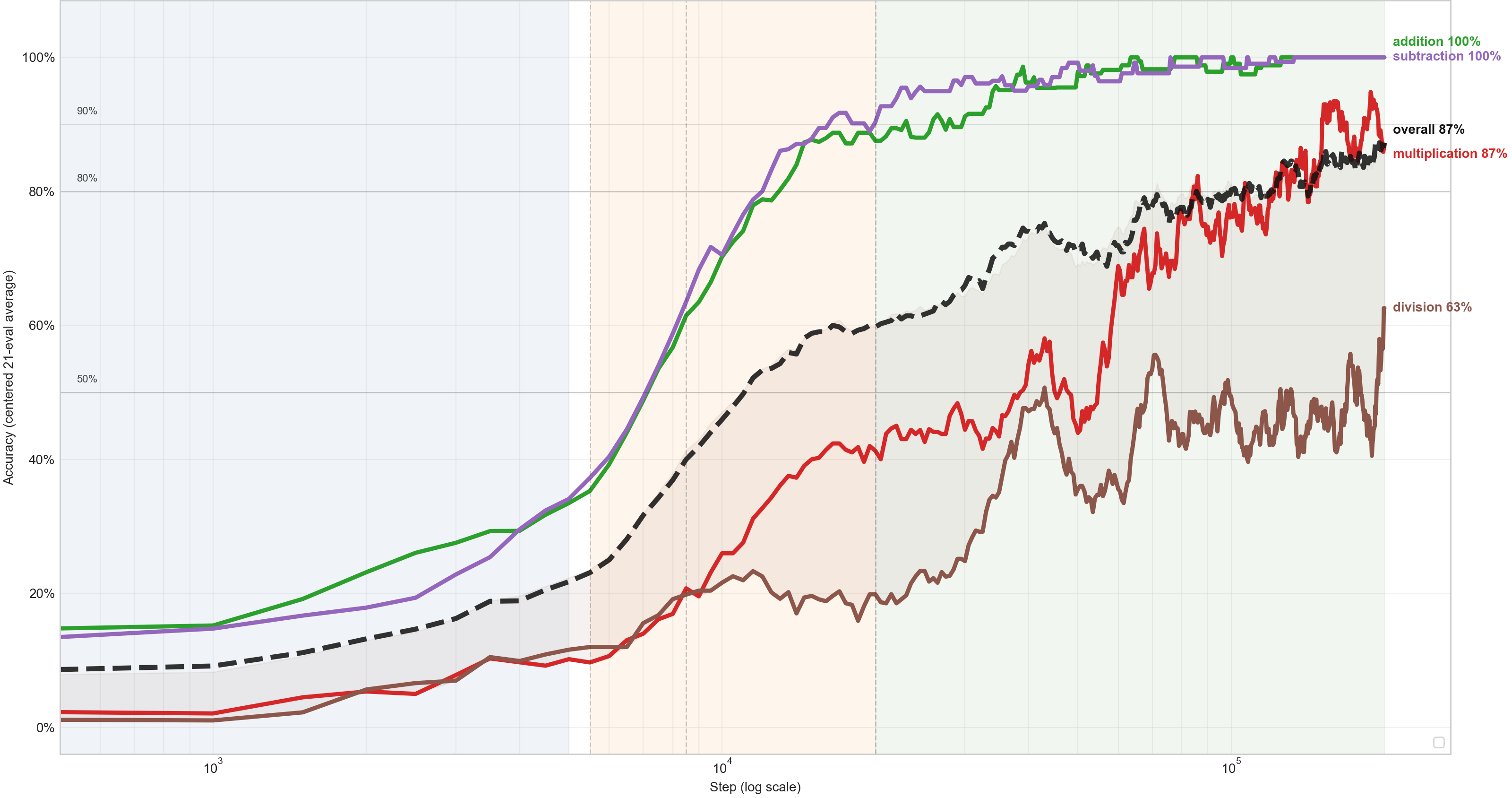}
\caption{Plot of the computation accuracy value for each type of arithmetic operation over the course of model training.}
\label{fig:figure4}
\end{figure}

Alongside the general monitoring of training, a detailed evaluation
of each arithmetic operation reveals the presence of
differing trajectories that reflect the level of ``ease'' of a given type of
operation. This is reflected in the number of training steps required
until the model reaches a high accuracy value on the operation concerned.
The multiplication operation demands more complex local arithmetic because
it involves summing partial products over small digits.
Meanwhile, the division operation requires determining the correct quotient
through a back-computation process that depends on the ability to
multiply. The higher computational complexity of these two
operations produces a learning trajectory that is gradual in nature.
As shown in Figure \ref{fig:figure4}, this process begins with
the mastery of the addition and subtraction operations, then followed by an
improvement in multiplication ability, and subsequently the development of
division ability as the number of training steps increases. At the end of the
training process, the model reaches an overall accuracy above 80\% on test
data that lies outside the provided training examples. This achievement
is considered high given the model's relatively small capacity, with a
scale of less than 100 million parameters.

\begin{figure}[h]
\centering
\includegraphics[width=\textwidth]{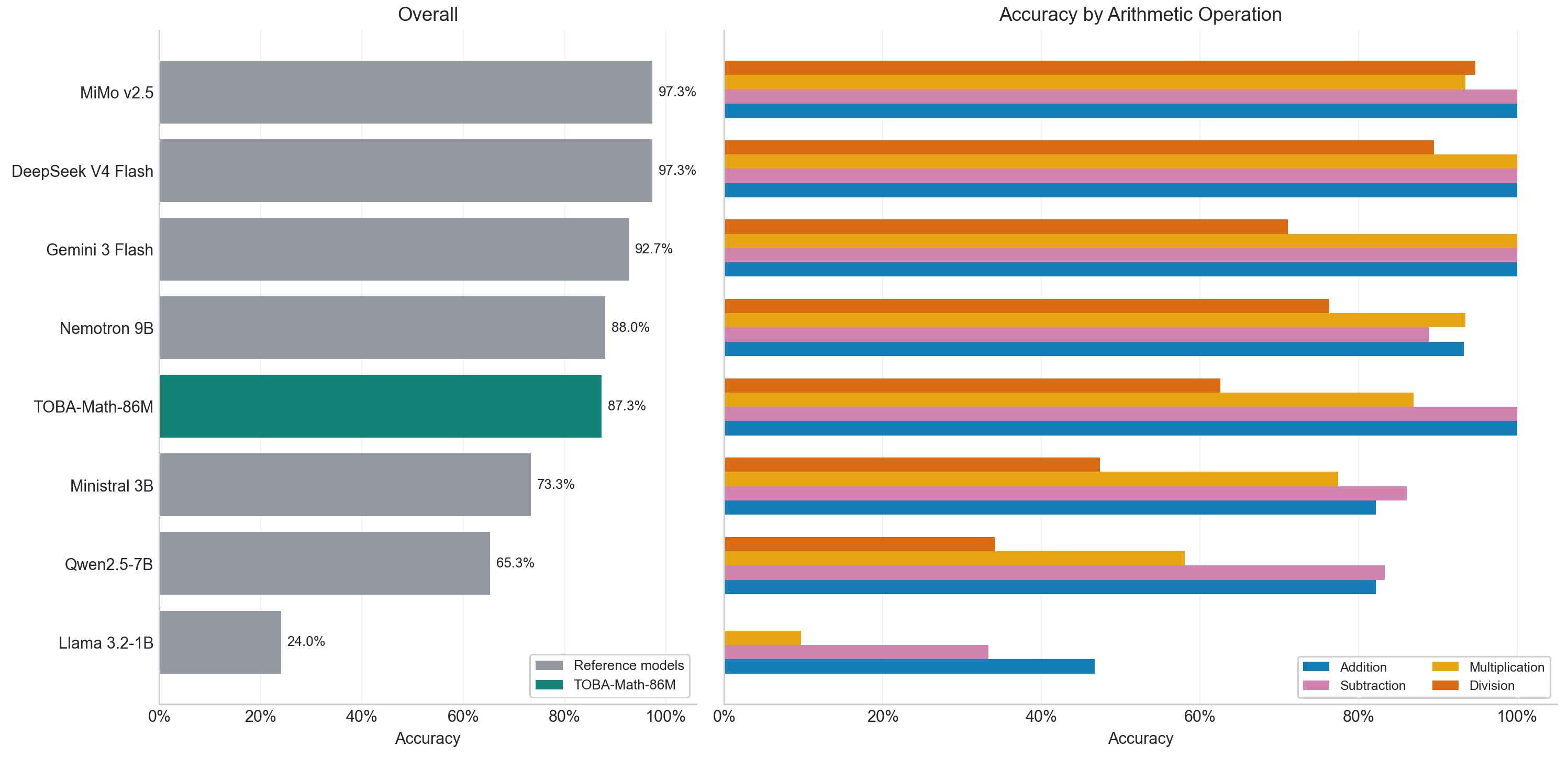}
\caption{\emph{Benchmarking} of basic arithmetic capability against various other large language models (LLMs).}
\label{fig:figure5}
\end{figure}

For comparison, we also perform an evaluation using the same test
data on several variants of language models with a larger parameter
scale. The comparison results show a consistent pattern, namely
that the multiplication and division operations tend to have a lower accuracy
level than addition and subtraction (Figure \ref{fig:figure5}).
Nonetheless, the model trained using the GASING method is able to
achieve better computation performance than several models
with a larger number of parameters. This finding confirms that
directed training that applies an effective mathematics pedagogy
can significantly improve the performance of a language model.

\begin{figure}[h]
\centering
\includegraphics[width=14cm]{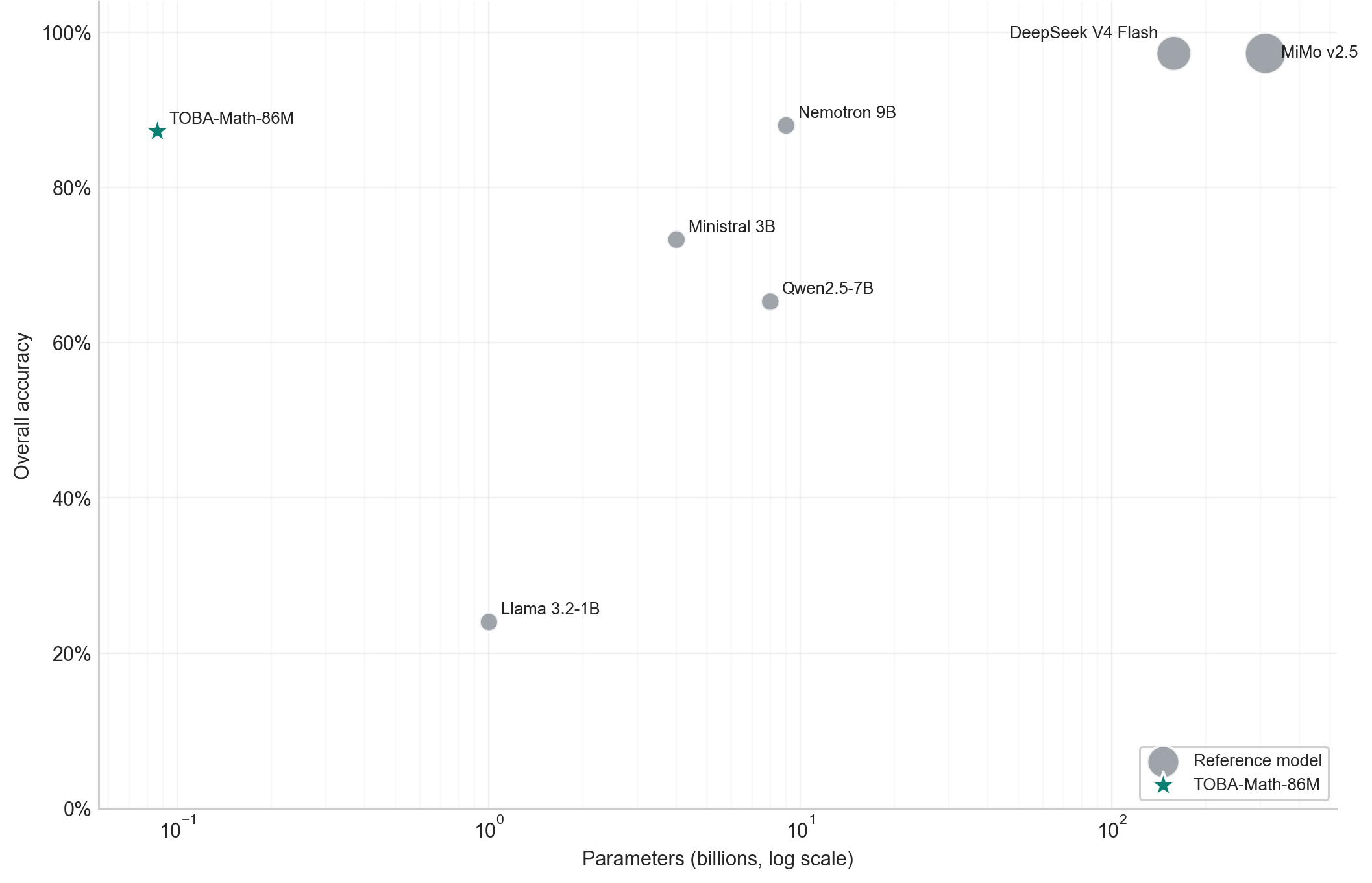}
\caption{Plot of the number of parameters \emph{vs.} computation accuracy value across various Transformer-based language models.
The trained model attains competitive performance despite a far more limited number of parameters.}
\label{fig:figure6}
\end{figure}

\section{Conclusion}
Language models designed to provide cognitive capacity within a linguistic framework have made considerable progress with the chain-of-thought (CoT) approach, which endows them with reasoning capability. Providing arithmetic reasoning to the reasoning process of language models is a further challenge addressed in this work. Natural human learning has put forward pedagogies for various fields of study, including arithmetic ability as a foundational part of the cognitive capacity for mathematical reasoning. The pedagogical approach to arithmetic with the GASING method has had a substantial impact on learners in Indonesia over the past several decades, and we attempt to apply it to a language model built upon the natural characteristics of Indonesian, namely TOBA-LM. Unlike conventional approaches to learning computation, the GASING method solves basic arithmetic problems by applying a ``left-to-right'' process, in which the decomposition of the computation begins from the position of the largest digit to the smallest. This distinctive feature is not merely a pedagogical variation but also carries computational consequences in the practice of language modeling. This is accomplished by transforming that arithmetic learning method into a computational procedure by constructing a textual dataset that serves as the source of supervision for running the language model training process.

The learning and training of this pedagogical method on a small-scale model with a unique tokenization such as TOBA-LM was examined in depth and yielded several interesting insights that offer potential for further investigation into how computational processes have an equivalence with natural human cognitive learning processes. Three phases with differing characteristics were found during arithmetic training on the language model. By periodically monitoring the cross-entropy loss, there is a first phase, namely the early phase, which we can interpret as the period during which the model learns the syntax and structure of language. This can be seen as a phase of synchronizing linguistic capacity with arithmetic capability, marked by the still relatively low arithmetic ability of the model. This phase is followed by a middle phase marked by a slowing of the loss decline. This indicates a learning process beginning to focus on a small number of tokens that constitute critical information. In this phase, the model's accuracy increases significantly from a value of less than 10\% to reaching about 60\%. In the final phase, the change in loss slows again, indicating a process of maturation of the representations that have been learned. Between the early and final phases lies a phase in which the learning process increases sharply and significantly, distinguishing the early arithmetic-learning phase from the process of sharpening arithmetic capability in the final phase.

Another point worth noting is the phenomenon of the emergence of an associative-memory aspect, a kind of mental arithmetic that in the GASING pedagogy is called ``mencongak'' (mental computation). The results of simple operations can appear directly without the need to run computation steps, no longer requiring an explicit procedure but relying instead on associative memory acquired through repeated exposure to data. These arithmetic-learning dynamics, together with the unique syllabic-based tokenization patterns of the TOBA-LM model, jointly facilitate more complex arithmetic reasoning capability. This opens an opportunity for further investigation into the relationship between linguistic cognitive capacity and the capacity for mental arithmetic.

Finally, the trained model reaches over 80\% accuracy on held-out problems and attains competitive performance against substantially larger language models, indicating that targeted, pedagogically grounded training can yield strong and economical arithmetic capability at small scale, the TOBA language model. This also opens innovative opportunities that can be explored further in order to economically enhance the arithmetic ability and capability of other models with a far larger number of parameters. This offers potential for the further development of various other language models with a gigantic number of parameters that may also possess superior mental-arithmetic capacity in the future.

\section*{Acknowledgement}
The authors thank Yohanes Surya for the discussions on the GASING arithmetic method, and Kevin Siringoringo for the TOBA-LM training pipelines presented in this report. All fault remains author's.

\bibliographystyle{plain}

\end{document}